\documentclass[3p, twocolumn, times]{elsarticle}

\usepackage{booktabs}

\usepackage[linesnumbered, lined, ruled, commentsnumbered, noend]{algorithm2e}
\let\oldnl\nl% Store \nl in \oldnl
\newcommand{\nonl}{\renewcommand{\nl}{\let\nl\oldnl}}% Remove line number for one line

\usepackage{amssymb}
\usepackage{amsthm}

\usepackage[colorlinks=true]{hyperref}
\usepackage{enumitem}

\journal{Expert Systems with Applications}

\begin{document}

\begin{frontmatter}

\title{TripMD: Driving patterns investigation via Motif Analysis}

\author[1]{Maria Inês Silva}
\ead{d20170088@novaims.unl.pt}
\author[2]{Roberto Henriques}
\ead{roberto@novaims.unl.pt}

\address{Nova Information Management School (NOVA IMS)\\
Campus de Campolide, Universidade Nova de Lisboa,\\
1070-312 Lisboa, Portugal}

\begin{abstract}

Processing driving data and investigating driving behavior has been receiving an increasing interest in the last decades, with applications ranging from car insurance pricing to policy making. A common strategy to analyze driving behavior is to study the maneuvers being performance by the driver. In this paper, we propose TripMD, a system that extracts the most relevant driving patterns from sensor recordings (such as acceleration) and provides a visualization that allows for an easy investigation. Additionally, we test our system using the UAH-DriveSet dataset, a publicly available naturalistic driving dataset. We show that (1) our system can extract a rich number of driving patterns from a single driver that are meaningful to understand driving behaviors and (2) our system can be used to identify the driving behavior of an unknown driver from a set of drivers whose behavior we know.

\end{abstract}


\end{frontmatter}

\section{Introduction}
\label{sec:intro}

In the last two decades, there has been a growing interest in analyzing driving data and understanding driving behavior, with researchers and practitioners finding new applications for this type of data. In the car insurance sector, measuring how a client drives is a cornerstone of the new usage-based insurance (UBI) schemes, which provide more custom pricing by taking into account driving behavior instead of external proxies such as sex and years of driving experience. In fleet management and fuel consumption optimization, studying the relationship between driving behavior and fuel consumption can improve driving performance and reduce costs. Regulators and policymakers can also leverage driving data to understand which factors are associated with accidents and improve road safety with better regulation. Analyzing how self-driving cars perform can help developers understand what is working correctly with the autonomous system and which areas need to be improved.

A common approach to get insights about driving performance from driving data is to analyze maneuvers. In fact, more and more authors use this type of analysis in their work \citep{velenis_optimality_2008, johnson_driving_2011, paefgen_driving_2012, saleh_driving_2017, skrjanc_evolving_2018, xie_driving_2018}. The rationale is that the set and frequency of the maneuvers performed during a trip and the way they are executed can provide relevant information about the driving behavior of the driver during the trip. So far, the driving data normally used in this task is high-frequency telematics (also called automobile sensor data), such as GPS location, velocity and acceleration, and video recordings.

In a previous article \citep{silva_finding_2020}, we argued that using time-series motifs detection algorithms to extract maneuvers from high-frequency telematics had two main benefits. First, it was more adaptable to small fluctuations in the data than previous methods. And second, it had the advantage of not requiring labels, which is extremely time-consuming to collect. We also noted that analyzing maneuvers through motif detection in telematics data is a promising area of research that is yet to be fully explored.

Recently, \citet{jain_masa_2019} proposed a general method to discover motifs in noisy time-series and, in one of their case studies, they concluded that their method was capable of identifying turn maneuvers from automobile sensor data. This work further validates our claim that motifs extracted from driving sensor data are highly related to the actual maneuvers performed.

In this paper, we expand the work done in \citep{silva_finding_2020} by proposing TripMD, a complete motif extraction and exploration system that is tailored for the task of analyzing maneuvers and driving behaviors. Other authors have looked into the task of maneuver detection using time-series motifs \citep{schwarz_time_2017, jain_masa_2019}. However, none of these works propose a full system that extracts motifs from automobile sensor data and summarizes the information in a space-efficient visualization.

Particularly, our main contributions are the following:

\setlist{nolistsep}
\begin{itemize}[noitemsep]
    \item We present TripMD, motif detection and summarization system that was designed to extract relevant driving patterns from a set of trips. It is the first system that not only extracts but also summarizes the main motifs of the provided trips, which allows for an easy investigation of the maneuvers being performed. TripMD expands our previous work by combining the motif extraction step with a motif clustering and summarization step.
    \item We present a novel representation method called Variable SAX (VSAX), which adapts the classical SAX representation from \cite{lin_experiencing_2007} to work with variable-length patterns. This new representation is what allows TripMD to capture maneuvers of variable lengths.

\end{itemize}

To evaluate the applicability of TripMD to real tasks, we use the UAH-DriveSet naturalistic driving dataset \citep{romera_need_2016} in two experiments. Firstly, we apply our system to the trips performed by a single driver and show that it is capable of extracting a rich set of driving patterns. We also show that these patterns can be used to distinguish between three different driving behaviors of the driver. Secondly, we demonstrate that, using the patterns extracted by TripMD, we are capable of identifying the driving behavior of an unknown driver from a group of drivers whose behavior we know. In other words, the association between driving patterns and driving behavior achieved with TripMD can generalize to unclassified drivers.

The rest of the paper is organized as follows. In Section \ref{sec:prelim}, we provide an overview of time-series motifs and motif detection algorithms, which will be helpful to understand TripMD. In Section \ref{sec:tripmd}, we describe TripMD in detail. Section \ref{sec:eval} is reserved for two experiments where we showcase our system and demonstrate its usefulness. And Section \ref{sec:conclusion} concludes this work and introduces some ideas for future work.

\section{Preliminaries}
\label{sec:prelim}

In simple terms, a time-series motif is a repeated pattern in the time-series that carries information about the underlying process that generated the time-series. Based on this general definition, there are two main ways of defining how relevant a repeated pattern is, namely based on support or based on similarity \citep{mueen_time_2014}. In the support-based definition, the most relevant pattern is the one with the highest number of repetitions, while, in the similarity-based definition, the most relevant pattern is the one with the most identical repetitions. Therefore, the support-based definition extracts more frequent patterns and the support-based definition extracts more similar patterns.

There are two additional constrains that a pattern needs to meet to be considered a time-series motif \citep{lin_finding_2002}. Firstly, two subsequences that belong to the same motif cannot overlap in time. This non-overlapping constraint is set to avoid trivial matchings. Secondly, two subsequences need to be at a distance smaller than a predefined radius $R$ to be considered a \textit{match} (and thus to belong to the same pattern).

Note that this second constraint is highly tailored to the use-case. On the one-hand, in most motif detection algorithms, the radius is a parameter that needs to be defined by the user. On the other hand, there are many distances that can be used to compute similarity between subsequences \citep{wang_experimental_2013, serra_empirical_2014} and the user must decide which distance is the most suitable to the specific use-case. 

The final constrain appears when the task is not to look for a single motif but to more than one motif. In this case, based on the motif definition, it is possible to order motifs based on their relevance and to extract the top-$k$ most important motifs, which are named the \textit{$k$-motifs}. However, any two motifs can only coexist in the list of \textit{$k$-motifs} if their centers (the subsequence that better represents the motif) have a distance higher than $2R$ \citep{lin_finding_2002}, where $R$ is the radius used to define the motif's matches.

In terms of the distance functions, the most used are the Euclidean distance \citep{das_rule_1998} and the \textit{Dynamic Time Warping} (DTW) distance \citep{berndt_using_1994}. The Euclidean distance performs an one-to-one comparison of single points from the same time location and, because of this, it is very efficient. However, the sequences being compared need to have the same size (or being padded at the end) and the distance is not robust to time-shifts, distortions or differences in phase. On the contrary, the DTW distance is capable of dealing with variable-length sequences and other misalignments by finding an optimal time mapping between the sequences that are being compared. However, this flexibility comes at the cost of efficiency, which is a major concern when analyzing time-series data.

Independently of the distance used, because motif discovery is a task that involves comparing all possible pairs of time-series subsequences, it is very computationally expensive. Thus, a lot of work in motif detection has been focused on making this search more efficient. The most used technique is to reduce the search space by converting the time-series into a low-dimensional representation where the true distance is approximately maintained. Then, we can prune motif candidates in this reduced space and search for the final motifs in the reduce group of candidates. The \textit{Symbolic Aggregate approXimation} \citep{lin_experiencing_2007}, or SAX, is the standard example of the this technique. It starts by braking the original time-series into fixed-sized sliding windows and then converting each window into a sequence of letters. 

Using the Matrix Profile (or MP) \citep{yeh_matrix_2016} is another commonly used strategy to speed-up the search for the motifs. The MP is a meta time-series that annotates the original time-series by providing the distance and the index of each fixed-size sequence's nearest neighbors, excluding trivial matches. Note that the size of the sequence is the only parameter of the method and the distance used is the Euclidean distance. There are many efficient and fast implementations of the MP, either approximate or exact, and after the MP is computed, extracting similarity-based motifs is trivial.

When working with telematic data, it is common to have data from several sensors such as the accelerometer and the velocimeter. Even in the case of the accelerometer, one still has two distinct time-series, namely the lateral and the longitudinal acceleration. Therefore, finding maneuvers in telematic data is a multidimensional problem and as such, we need to apply techniques for detecting multidimensional motifs.

In their work, \citet{tanaka_discovery_2005} suggested to apply a dimensionality reduction technique in order to reduce the multidimensional time-series into a single dimension, which would simplify the problem back to the one-dimensional case. This is a smart approach as it can easily leverage all the existing motif detection algorithms. However, it has the drawback of information loss. If the time-series data contains relevant information in more than one dimension at the same time (which is our case when using acceleration data), then we won't be able to capture all the relevant motifs with this approach.

Another technique widely used in the multidimensional setting is to apply an one-dimensional motif detection algorithm in each dimension independently and then look for co-occurrences to extract the multidimensional motifs \citep{minnen_detecting_2007, vahdatpour_toward_2009, balasubramanian_discovering_2016, liu_multi-dimensional_2017}. This setup is much more accurate (since there is no information loss) and it is more flexible (since the search for co-occurrences can be done with an allowance for asynchronous motifs and the rejection of uninformative dimensions). However, this setup is more computationally expensive.

Instead of working on a two step approach, other authors search for the multidimensional motifs directly by concatenating all the dimensions. For instance, \citet{minnen_detecting_2007} compute the SAX representation for each dimension and concatenate their strings, while in \citep{yeh_matrix_2017} the authors define a new Matrix Profile, the k-dimensional MP, that encodes the distance to each sequence's closest neighbor, taking into account k dimensions.

When the goal is to analyze maneuvers, one needs to be able to extract variable-length motifs. In other words, because the same maneuver does not always take the exact same time, it is important to have flexible methods that can extract motifs of different lengths. Even though fixed-length motifs have been the most explored so far, there are some algorithms for the variable-length case.

Most authors propose to apply a fixed-length algorithm in a range of window sizes and then choose the most representative motifs based on their ranking scheme. The work of \citet{nunthanid_parameter-free_2012} and \citet{gao_exploring_2018} are two examples of this approach. Note, however, that this approach does not work for maneuvers since all the subsequences that belong to a given motif have the same size. If we have a trip with three right-turn maneuvers that have slightly different durations, these algorithms would not be able to identify these three maneuvers as a motif.

Lin's grammar-based method \citep{lin_finding_2010} and Tanaka's EMD algorithm \citep{tanaka_discovery_2005} take a different approach. They adapt the sequences' representation in the low-dimensional space in order to take into consideration variable-length patterns. However, this adaption leads to algorithms that are not exact, which means that there is no guarantee that the method can find all the variable-length motifs in a given time-series.

\section{TripMD}
\label{sec:tripmd}

TripMD is a system for extracting and analyzing maneuvers from trips performed by a single driver. To achieve this, we needed algorithms that worked on multi-dimensional time-series and, at the same time, were able to extract and analyze variable-length patterns. Having this in mind, our solution has the following two components:

\begin{enumerate}[noitemsep]
    \item A motif extraction algorithm inspired by the algorithm created by \citet{tanaka_discovery_2005}, which was tailored and tuned for the maneuver detection use-case. It includes a discrete variable-length representation (variable SAX) based on the widely used SAX \citep{lin_experiencing_2007} and an iterative pattern matching process that extracts motifs in multiple dimensions. 
    \item A motif clustering and visualization tool based on the Self-Organizing Map model that extracts the most relevant motif patterns and permits the user to quickly analyze them.
\end{enumerate}

In our motif extraction algorithm, we use the support-based definition. In other words, for a certain variable SAX (VSAX) pattern, its motif is the largest group of non-overlapping variable-length subsequences with that VSAX representation and in which all the subsequences have a distance lower than a predefined radius $R$ to the motif's center. Because we are working with variable-length motifs, we use the \textit{Dynamic Time Warping} (DTW) distance \citep{berndt_using_1994} to measure similarity between two multi-dimensional subsequences.

In the following subsections, we'll go through each component in more detail.

\subsection{Motif extraction}

The motif extraction step of TripMD receives a list of trip recordings $T$ and some parameters and returns the list $M$ of all the motifs in those trips that meet our support-based definition. 

\begin{algorithm}[ht!]
\small
\label{alg:motif_extraction}
\caption{Motif extraction}
\DontPrintSemicolon

\KwIn{$T$: List of multidimensional time-series representing the trips' recordings\newline
      $l_{size} \in \mathbb{Z}^+$: Default letter size\newline
      $w_{min}\in \mathbb{Z}^+$: Minimum pattern size\newline
      $R \in \mathbb{R}^+$: Motif radius}
\KwOut{$M$: List of all motifs.}
\BlankLine

    $V \gets \textsc{GetVsaxSequence}(T, l_{size})$ \;
    $M \gets \emptyset$\;
    $w \gets w_{min}$\;
    \Repeat{$|W_{unique}^w| = |W^w|$ (i.e., no more repeating patterns)}{
        $W^w \gets$ sequence of VSAX words of size $w$\;
        $W_{unique}^w \gets$ list of unique word patterns\;
        \ForAll{$pattern \in W_{unique}^w$}{
            $m \gets \textsc{GetMotif}(pattern, R, W^w, T)$\;
            \uIf{number of motif members of $m > 1$}{
                $M \gets M \cup \{m\}$\;
              }
        }
        $w \gets w + 1$
    }
\end{algorithm}

Algorithm \ref{alg:motif_extraction} presents the motif extraction algorithm. First, we compute the list of sequences $V$ that contains the VSAX representation of $T$. This step is further described in Section \ref{subsec:vsax}, but in short, each element of $T$ is a sequence of symbols that summarize a particular time-series of $T$. 

Next, for each pattern size $w$, we extract a list of all words in $V$ for that size ($W^w$). A word in $W^w$ is a sequence of $w$ VSAX symbols that correspond to a variable-length subsequence in $T$.

Then, for each unique word in $W^w$, if it exists, we extract its corresponding motif using the algorithm described in Section \ref{subsec:motif-search}. We add all the motifs found to the list $M$ and the loop ends when no more motifs of a certain pattern size exist. 

\subsubsection{Variable SAX}
\label{subsec:vsax}

Variable SAX (VSAX) is a time-series discretization method that transforms a time-series into a sequence of symbols that captures the general behavior of the original time-series. It serves two main purposes. Firstly, by providing a discretization of the time-series, it allows for a more efficient motif search and, at the same time, reduces the impact of small levels of noise. Secondly, it is capable of splitting the time-series into subsequences of variable lengths depending on the underlying behavior of the time-series, which allows a simple pattern matching algorithm to find variable-length motifs. 

Algorithm \ref{alg:vsax} lists the main steps of the VSAX algorithm. For simplicity, we present the case where the trips are represented as an one-dimensional time-series. The application of the algorithm to the multi-dimensional case is discussed in the last paragraph of the subsection.

\begin{algorithm}[ht!]
\small
\label{alg:vsax}
\caption{Variable SAX (VSAX)}
\DontPrintSemicolon

\KwIn{$T$: List of one-dimensional time-series\newline
      $l_{size} \in \mathbb{Z}^+$: Default letter size}
      
\KwOut{$V$:  list of VSAX letter sequences, one sequence per trip}

\nonl Let $\alpha_i$, be the symbol that represents the time-series domain $[b_{i}, b_{i+1}[$, where $1 \leq i \leq 5$.  \;
\BlankLine

    \textbf{Function} $\textsc{GetVsaxSequence}(T, l_{size})$: \;
        $V \gets \emptyset$ \;
        $\{b_i\}_{1 \leq i \leq 6} \gets \textsc{ComputeVsaxBreakpoints}(T)$ \;
        
        \ForEach{time-series $ts$ of $T$}{
            $V_{ts} \gets \emptyset$ \;
            $ts_{size} \gets$ size of the trip's time-series $ts$ \;
            \For{$k \leftarrow 1$ \KwTo $ts_{size} - l_{size}$}{
                $tss_k \gets ts[k:k+l_{size}]$\;
                $paa_k \gets \frac{1}{l_{size}} \sum tss_k[i]$\;
                $c_k \gets \alpha_i$ s.t. $paa_k \in [b_{i}, b_{i+1}[ $\;  
                \eIf{$|V_{ts}| = 0$ or $c_k \neq c_{k-1}$}{
                    $letter_k \gets (c_k, tss_k)$\;
                    $V_{ts} \gets V_{ts} \cup \{letter_k\}$
                }{
                    $tss_k = tss_{k-1} \cup tss_k$\;
                    $letter_{k} \gets (c_k, tss_k)$\;
                    $V_{ts} \gets (V_{ts} \setminus \{letter_{k-1}\}) \cup \{letter_k\}$
                }
            }
            $V \gets V \cup \{V_{ts}\}$
        }
    \textbf{return} $V$
\end{algorithm}

\begin{figure}[ht!]
    \centering
    \includegraphics[width=0.95\linewidth]{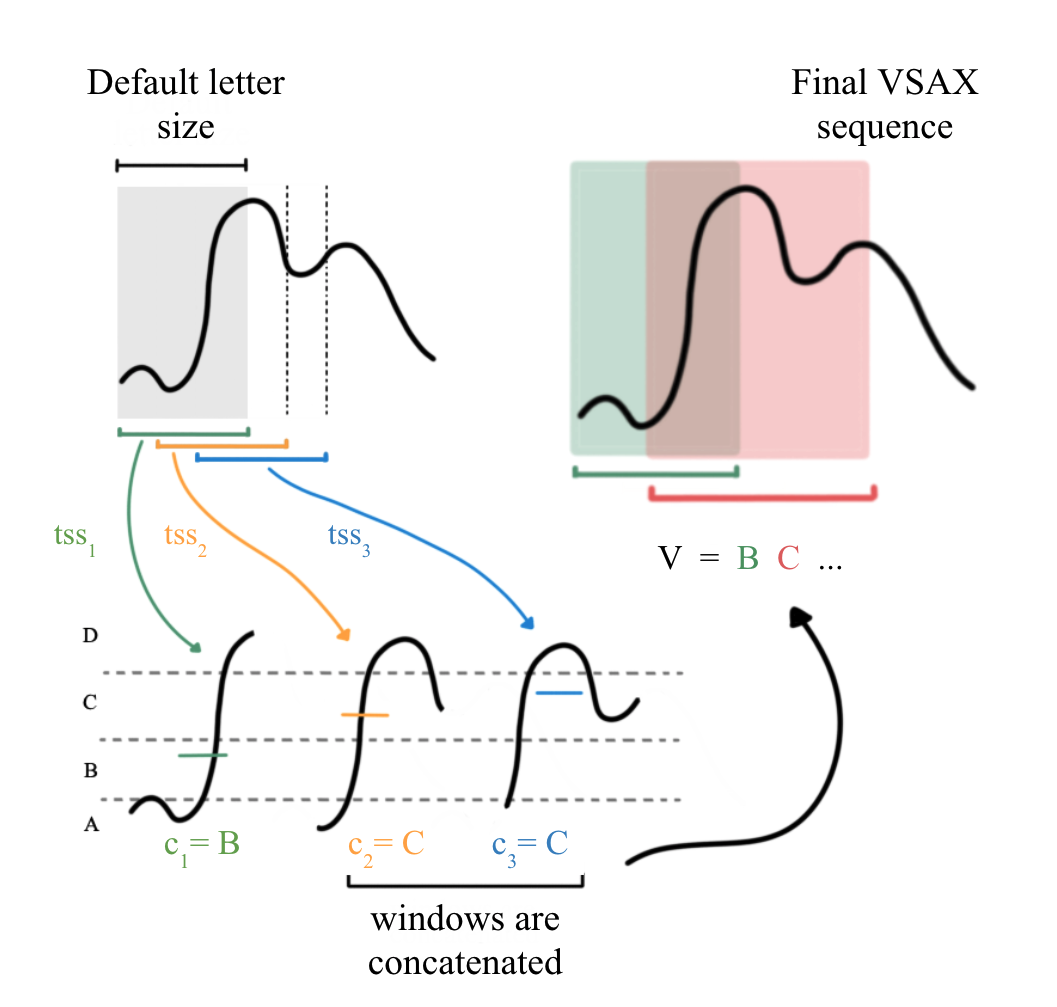}
    \caption{Simple example of the Variable SAX representation process.}
   \label{fig:vsax}
\end{figure}

To better explain how the algorithm works, we provide a example in Figure \ref{fig:vsax} with a single one-dimensional time-series.

Initially, the time-series is split into fixed-length sliding windows ($tss_k$). The length of the window is one of the parameters of VSAX, the default letter size. Then, the values in each sliding window are averaged to obtain a discrete value for that window ($paa_k$), which in turn is converted to a symbol ($c_k$) based on a predefined segmentation of the time-series domain.

After obtaining the symbol $c_k$, the pruning phase concatenates checks whether $tss_k$ should be concatenated to the previous window $tss_{k-1}$ (line 14 in Algorithm \ref{alg:vsax}). The rationale is that if two consecutive windows have similar behaviors (which translates into being transformed into the same symbol), then they should be a single window and be considered together when searching for motifs. Thus, in the end, we have a sequence of symbols $V$ that map to variable-length subsequences of the original time-series and that encode information about the general behavior of the subsequences.

The segmentation of the time-series domain is similar to the way it is done in SAX \citep{lin_experiencing_2007}. Break-points are determined based on the time-series' values and these break-points define regions in the time-series domain that map to specific symbols. Algorithm \ref{alg:vsax_breakpoints} details how these break-points are computed in TripMD, assuming the input is an one-dimensional time-series of trips recordings.

\begin{algorithm}[ht!]
\small
\label{alg:vsax_breakpoints}
\caption{Variable SAX break-points computation}
\DontPrintSemicolon

\KwIn{$T$: List of one-dimensional time-series}
\KwOut{$\{b_i\}_{1 \leq i \leq 6}$: VSAX break-points for $T$}
\BlankLine

    \textbf{Function} $\textsc{ComputeVsaxBreakpoints}(T)$:\;
        $joinedT \gets \emptyset$\;
        \lForEach{time-series $ts$ of $T$}{$joinedT \gets joinedT \cup ts$}
        $b_1 \gets - \infty$\;
        $b_2 \gets \textsc{Percentile}(T, 5)$\;
        $b_3 \gets \textsc{Percentile}(T, 15)$\;
        $b_4 \gets \textsc{Percentile}(T, 85)$\;
        $b_5 \gets \textsc{Percentile}(T, 95)$\;
        $b_6 \gets + \infty$\;
    \textbf{return} $\{b_1, b_2, b_3, b_4, b_5, b_6\}$
\end{algorithm}

In the original SAX representation \citep{lin_experiencing_2007}, break-points are defined so that all regions have equal probability under a Gaussian distribution. However, VSAX uses specific percentiles of the time-series' values to define five regions, namely the 5th, 15th, 85th and 95th percentiles. In general, a driver spends less time performing maneuvers than he does not performing any maneuver and, thus, defining break-points that evenly distribute time-series' values among the regions does not lead to good results in the maneuver detection task. Additionally, since the percentiles are computed over all the trips, any two windows with the same symbol will be guaranteed to be in the same domain region. This is another change compared to SAX, where the break-points are computed independently for each window.

Finally, for multi-dimensional time-series, VSAX can be applied separately to each one-dimensional time-series and then concatenate the resulting symbols in a tuple. For instance, a subsequence of a two-dimensional time-series would be mapped to a tuple of two symbols, one for each dimension. Note however that in the multi-dimensional case, the pruning phase is applied in all the dimensions at the same time. In other words, two consecutive subsequences are only merged if they have the same symbols in all the dimensions. Thus, in this case, each VSAX symbol corresponds to a single variable-length multi-dimensional subsequence of the original time-series.

\subsubsection{Motif search}
\label{subsec:motif-search}

The motif search is an iterative process that extracts the motifs of all possible sizes from a VSAX sequence. It was inspired by the motif detection algorithm proposed by \citet{tanaka_discovery_2005}. At each iteration, it discovers all the motifs with a certain number of VSAX symbols (the pattern size) and then moves to next iteration by increasing the pattern size by one. The minimum pattern size is a parameter of the method and the iteration stops when no more motifs with the current pattern size can be found. Algorithm \ref{alg:get_motif} lists the main steps to discover the motif associated with a pattern word, if it exists.

\begin{algorithm}[ht!]
\small
\label{alg:get_motif}
\caption{Motif discovery from a word pattern}
\DontPrintSemicolon

\KwIn{
     $pattern$: pattern word\newline
     $R \in \mathbb{R}^+$: Motif radius\newline
     $W^w$: sequence of VSAX words of size $w$\newline
     $T$: List of multidimensional time-series representing the trips' recordings}
\KwOut{$(m_{center}, m_{members})$: center and members of the final motif, it they exist}
\BlankLine
\textbf{Function} $\textsc{GetMotif}(pattern, R, W^w, T)$:\;
    $m_{candidates} \gets$ all subsequences with the word $pattern$\;
    $D \gets$ matrix of the DTW distances of all subsequence pairs of $m_{candidates}$\;
    $count_{max} \gets 0$\;
    $m_{mean} \gets \infty$\;
    $m_{center} \gets \emptyset$\;
    $m_{members} \gets \emptyset$\;
    \ForEach{candidate $c$ of $m_{candidates}$}{
        $c_{members} \gets \{s \in m_{candidates} \mid distance(c, s) \leq R \land c, s$ do not overlap$\}$\;
        $count_{new} \gets |c_{members}|$\;
        \uIf{$count_{new} > count_{max}$}{
            $count_{max} \gets count_{new}$\;
            $m_{mean} \gets$ Mean distance between $c_{members}$ and $c$\;
            $m_{center} \gets c$\;
            $m_{members} \gets c_{members}$\;
        }
        \uElseIf{$count_{new} = count_{max}$}{
            $c_{mean} \gets$ Mean distance between $c_{members}$ and $c$\;
            \If{$c_{mean} > m_{mean}$ }{
                $count_{max} \gets count_{new}$\;
                $m_{mean} \gets c_{mean}$\;
                $m_{center} \gets c$\;
                $m_{members} \gets c_{members}$\;
            }
        }
}
\textbf{return} $(m_{center}, m_{members})$
\end{algorithm}

\begin{figure}[ht!]
    \centering
    \includegraphics[width=0.98\linewidth]{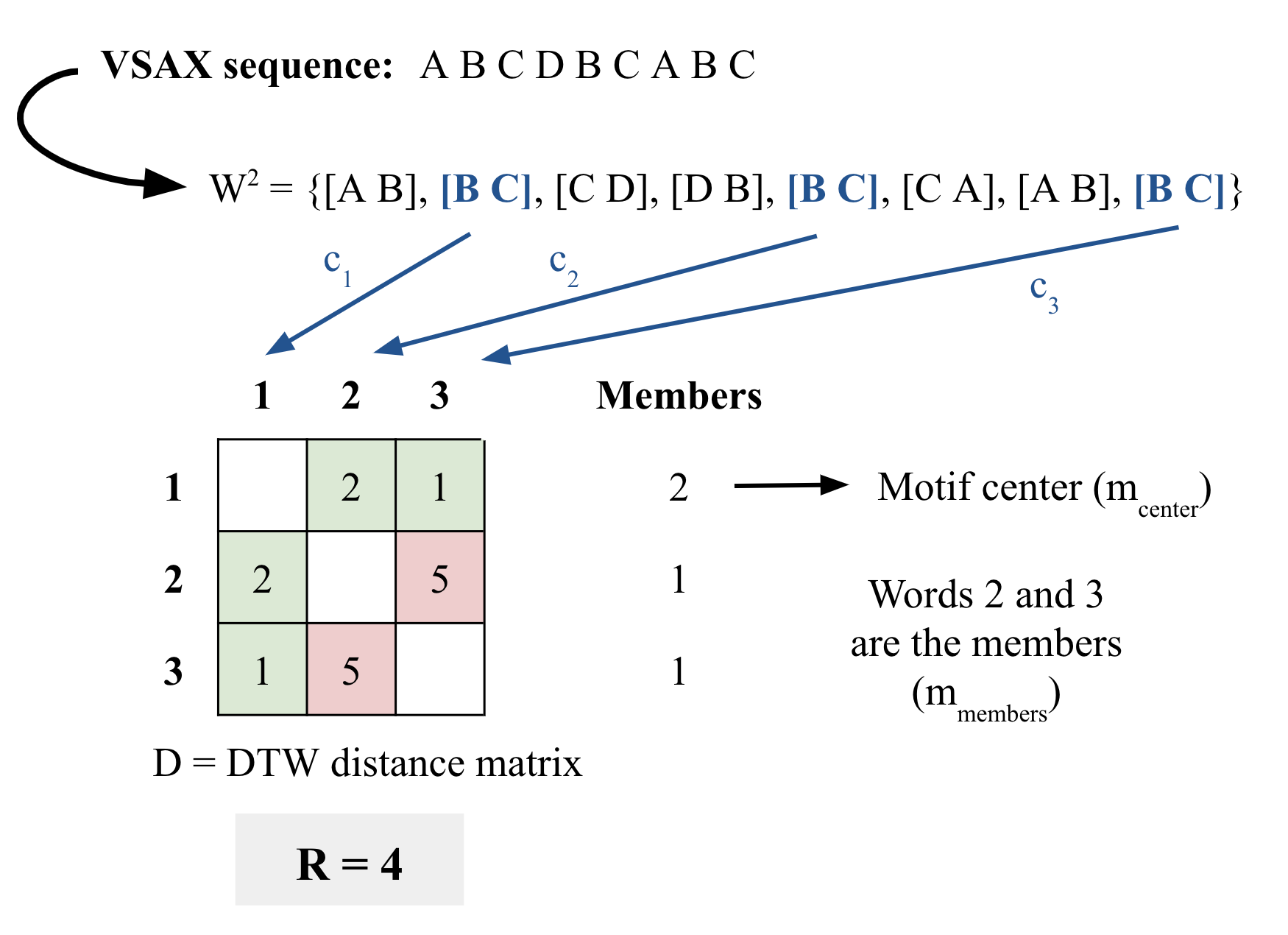}
    \caption{Simple example of the motif search process for a single pattern word \textit{BC}.}
   \label{fig:motif-search}
\end{figure}

To further illustrate the algorithm, Figure \ref{fig:motif-search} provides a concrete example. In this case, the pattern word is \emph{BC} and the motif radius $R = 4$. Initially, we extract all the subsequences with that same pattern to make the pool of the motif's center candidates $c_{candidates}$. In the example, $c_{candidates}$ has 3 subsequences.

From $c_{candidates}$, we compute the matrix of DTW distances for all the candidates ($D$), which allows us to extract the motif members of a center candidate.

Then, we loop through all subsequences in $c_{candidates}$ and get the corresponding motif members, $m_{candidates}$,  assuming that the subsequence $c$ is the center of the motif. Concretely, for a given center candidate $c$, all the non-overlapping candidates that are within $R$ from the initial candidate are extracted and stored in $c_{members}$ (line 9 of Algorithm \ref{alg:get_motif}).

The final motif is defined to be the set of subsequences with the most members ($m_{members}$), and the motif's center is the original candidate that generated that set of members ($m_{center}$). If no set of more than one member is found, then the motif for that specific pattern does not exist and both $m_{members}$ and $m_{center}$ are empty. In the example, the final motif exists and has the first subsequence as its center and the remaining two subsequences as its members. 

\subsection{Motif summarization}

In a previous work \citep{silva_exploring_2020}, we proposed a new dimensionality reduction method to summarize and explore the outputs of any motif detection algorithm. The method, called DTW-SOM, is a vanilla Self-Organizing Map \citep{kohonen_self-organizing_2001} with some adaptions to work with time-series motifs. It receives a list of variable-length multi-dimensional motifs and produces a clustering of the motifs' centers and a visualization of the results that is space-efficient.

TripMD leverages the DTW-SOM algorithm to group all the motifs found by the motif search process and to provide a visual summary of the most relevant motifs in the trips under analysis. By summarizing the extracted motifs, the user can quickly analyze the main patterns that are extracted and can better interpret the maneuvers being performed.

However, there is an important step before applying this method. DTW-SOM includes two initialization routines, a random initialization, in which the DTW-SOM network is initialized with a random sample of the motifs, and an anchor initialization, in which the user provides the set of motifs to used for initialization. Since previous experiments indicated that the anchor initialization was more stable than the random initialization \citep{silva_exploring_2020}, TripMD uses the anchor initialization. This means that it has to include a motif pruning step (Algorithm \ref{alg:pruning}) that computes the most relevant motifs, which are used as the anchors.

\begin{algorithm}[ht!]
\small
\label{alg:pruning}
\caption{Motif pruning}
\DontPrintSemicolon

\KwIn{
     $M$: list of motifs\newline
     $R \in \mathbb{R}^+$: Motif radius}
\KwOut{$P$: Pruned list of motifs, ordered by MDL score, from lowest to highest}
\BlankLine

$scores \gets \emptyset$\;
\ForEach{motif $m$ of $M$}{
    $mdl_m \gets \textsc{MdlScore}(m)$\;
    $scores \gets scores \cup \{mdl_m\}$
}
$M^* \gets \textsc{Sort}(M, by=scores)$\;
$P \gets \{M^*[1]\}$\;
\ForEach{motif $m$ of $M^*$}{
    $distances \gets \emptyset$\;
    \ForEach{motif $p$ of $P$}{
        $d \gets$ DTW distance between the centers of $p$ and $m$\;
        $distances \gets distances \cup \{d\}$
    }
    \If{$d \leq 2R \ \forall d \in distances$}{
        $P \gets P \cup \{m\}$
    }
}
\end{algorithm}

The pruning routine used in TripMD (Algorithm \ref{alg:pruning}) is based on the definition of $k$-motifs and the non-overlapping requirement discussed in Section \ref{sec:prelim}. Given a natural ordering of all the extracted motifs, the $k$-motif is the highest ranking motif whose center has a distance higher than $2R$ to each of the $j$-motifs' centers, for $1 \leq j \leq k-1$. Thus, the first pruned motif is the  first element of the sorted list of motifs ($M^*[1]$). Then, the next subsequence motif to be added to the list of pruned motifs $P$ is the first motif in the sorted list $M^*$ that has a distance higher than $2R$ to the motif in $P$. And so forth for the remaining motifs in $M^*$.

In TripMD (line 5 of Algorithm \ref{alg:pruning}), the ordering is defined by the MDL cost proposed by \citet{tanaka_discovery_2005}. This score is based on the Minimum Description Length (MDL) principle \citep{rissanen_stochastic_1998} which states that the \textit{"best model to describe a set of data is the model which minimizes the description length of the entire data set"} \citep{tanaka_discovery_2005}. In other words, by using the MDL principle, we are ranking the motifs based on their capacity to compress the original time-series data. In this case, the lower the MDL score, the more relevant the motif is. After pruning, the lowest-scored motifs ($P$) are used to initialize the DTW-SOM and all the motifs are fed into the algorithm.

TripMD also imposes a constrain on the distance computation of DTW-SOM. The original SOM algorithm \citep{kohonen_self-organizing_2001} uses the Euclidean distance as its distance metric. However, because DTW-SOM needs to work with variable-length subsequences, it uses the DTW distance instead.

As presented in Section \ref{sec:prelim}, DTW allows a more flexible comparison of two subsequences by finding the optimal time mapping between them. This flexible mapping is also called time warping. If no constraint is provided, DTW searches are possible mappings between the two subsequences to find the one with the lowest distance. Alternatively, DTW can limit the maximum time wrapping allowed. In this case, DTW cannot map two time-steps that are farther in time than the maximum warping window. Thus, because the search is constrained, the final mapping will be sub-optimal, and the resulting distance will be higher or equal to the one obtained with an unconstrained version of DTW.

\subsection{Parameter analysis}

So far, TripMD seems to have some parameters that a user needs to set beforehand. VSAX has the default letter size, the motif search has the radius $R$ to define the motifs and the minimum pattern size that initializes the search, and DTW-SOM has the number of training epochs and the maximum warping window. However, TripMD has default values for these parameters, namely:

\begin{itemize}[noitemsep]
    \item \textbf{Default letter size:} 1 second
    \item \textbf{Minimum pattern size:} 3 VSAX letters
    \item \textbf{Motif radius:} 0.5th percentile of the distance between all pairs of 3 second subsequences
    \item \textbf{Number of epochs for DTW-SOM:} 20
    \item \textbf{Maximum warping for DTW-SOM:} VSAX's default letter size (or 1 second)
\end{itemize}

Using the default parameters, the only parameter that the user must provide is the frequency in Hertz of the input time-series, which is trivial. If the user has some particularity in his dataset that makes the default parameters unreasonable, there's always the possibility of overriding the defaults provided by the TripMD.

To better understand how these parameters influence our method, we use a small toy example and test it against varying parameter values. We build a toy example by picking two right turns from a real driver and concatenating them with some random noise. The noise was obtained using an uniform distribution $U(-0.005, 0.005)$. Thus, we obtain the lateral and longitudinal acceleration of a small trip with two right turns connected by random noise.

\begin{figure}[ht!]
    \centering
    \includegraphics[width=0.95\linewidth]{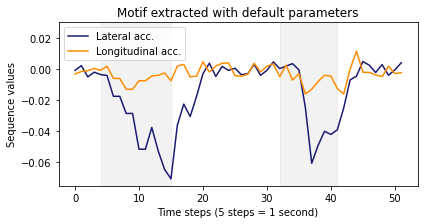}
    \caption{Motif extracted from the toy example using the default parameters. It includes the lateral and longitudinal acceleration values of the entire toy example. The two subsequences that belong to the motif are identified by the shaded area.}
   \label{fig:toy-motif-default}
\end{figure}

Since the data has a frequency of 5 Hz, the default settings of TripMD lead to a default letter size of 5 and a minimum pattern size of 3. We set the default motif radius to 0.0684 as this was the value estimated by our method using all the trips of the driver from which we picked the two right turns of the toy example. With these default parameters, we run the motif extraction component of TripMD and obtained two motifs. Figure \ref{fig:toy-motif-default} shows one of the motifs extracted, which includes the two subsequences of the original right turns, confirming that the default parameters work well for the toy example.

In TripMD, the default letter size controls the degree of detail used to define driving patterns. This parameter sets the size of the sliding windows used in the VSAX discretization. Thus, if the parameter is too large, the sliding windows will be too large as well, and TripMD will lose the detail necessary to identify relevant maneuvers. On the other hand, if the parameter is too small, the sliding windows will be too short as well, and TripMD will be too sensitive to noise. We define the default value to be one second because in the dataset we had available one second was the right duration to identify simple changes in acceleration that are related to actual maneuvers. However, this should be further tested in other datasets.

\begin{figure}[ht!]
    \centering
    \includegraphics[width=0.95\linewidth]{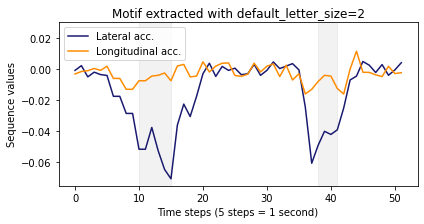}
    \includegraphics[width=0.95\linewidth]{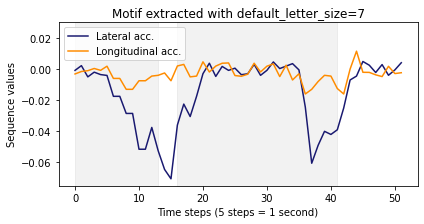}
    \caption{Motifs extracted from the toy example with varying default letter sizes. The subsequences that belong to the motifs are identified by the shaded area.}
   \label{fig:toy-motif-letter-size}
\end{figure}

Figure \ref{fig:toy-motif-letter-size} shows the motifs extracted from the toy example using two default letter sizes, one small (size 2) and one large (size 7). As explained in the previous paragraph, when the window size is small, the motif extracted is also small, and we can never extract the entire turn maneuvers. When the window size is large, the motif extracted expands beyond the subsequences associated with the turns.

The minimum pattern size corresponds to the minimum number of VSAX letters used to build motifs. Setting the default to three letters means that all the motifs will correspond to subsequences with two changes in acceleration. For instance, a simple turn maneuver should include two changes in the lateral acceleration: from zero to absolute high and back again to zero. Three letters are the minimum number of changes to have a meaningful maneuver.

If the minimum pattern size is set to a lower value, TripMD will still extract motifs with more letters. However, the search will take longer to finish. On the other hand, if the minimum pattern size is higher than three, TripMD will not find simple maneuvers such as a right turn or a brake.

\begin{figure}[ht!]
    \centering
    \includegraphics[width=0.95\linewidth]{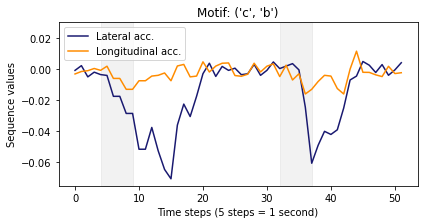}
    \includegraphics[width=0.95\linewidth]{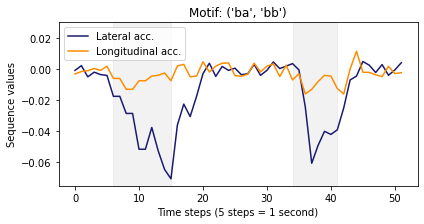}
    \caption{Motifs extracted from the toy example using a minimum pattern size 1. One of the motifs corresponds to a pattern size of 1 and the other corresponds to a pattern size of 2. The subsequences that belong to the motifs are identified by the shaded area.}
   \label{fig:toy-motif-pattern-size}
\end{figure}

As an example, Figure \ref{fig:toy-motif-pattern-size} shows two motifs extracted with a minimum pattern size 1. Both motifs are parts of the initial motif extracted with the default parameters. Because the motif search algorithms needs to search through these smaller sized motifs, the computation takes longer and does not arrive at any extra meaningful maneuvers.

The motif radius parameter, $R$, limits the subsequences that are considered to be a motif. Even if two subsequences have the same pattern word, they will only belong to the same motif if their DTW distance is lower than $R$. Thus, if $R$ is too small, very few motifs will be found. And if $R$ is too large, clearly different subsequences will be added to the same motif, which will result in poor results.

When estimating this parameter, we assume that motifs are rare and that most pairs of subsequences in a trip will not be the same maneuver. Thus, we use a small percentile on the distribution of the distance between random pairs of subsequences to guarantee that we use a $R$ large enough to find motifs without the risk of grouping different maneuvers in the same motif.

\begin{figure}[ht!]
    \centering
    \includegraphics[width=0.95\linewidth]{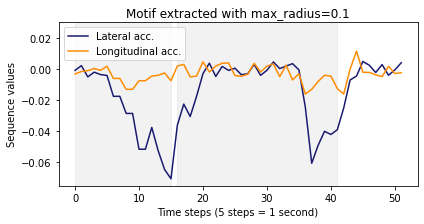}
    \caption{Motif extracted from the toy example using a motif radius of 0.1. The subsequences that belong to the motif are identified by the shaded area.}
   \label{fig:toy-motif-radius}
\end{figure}

Going back to the toy example, if the radius is set to value lower than 0.04, no motifs are extracted. However, setting $R$ higher or equal to 0.05 allows us to extract the motif with the two turn maneuvers. The difference, however, between using a radius of 0.05 and a radius of 0.07 is the extra patterns that are extracted when one increases $R$. As an example, Figure \ref{fig:toy-motif-radius} shows a motif extracted with a motif radius of 0.1.

The final two parameters are related to the DTW-SOM method. The number of epochs sets the number of training iterations for the SOM. The higher the number of epochs, the longer the model will train (and the longer the full run of TripMD will be). However, if the number of epochs is too small, the final summary of the motifs will be poor. Since we are using the anchor initialization, we don't need as many epochs for the SOM training to converge. From our experiments, the default of 20 epochs is sufficient for a proper summarization. 

As for the maximum warping, it limits the time-warping allowed in the DTW distance computation. If we decrease this parameter, the maximum warping window is small, and the distance between the two subsequences will be closer to their Euclidean distance. Alternatively, if we allow for more warping, the clustering of the motifs will be more flexible and allow more misaligned sequences to be grouped in the same SOM unit.

There are two reasons for limiting the warping window. Firstly, since we are reducing the search for an optimal mapping, the computation is faster. Secondly, if the warping window is not constrained, motifs with a higher degree of time misalignment will be grouped in the same SOM unit, and the final motif summarization will be distorted.

\section{Evaluation and discussion}
\label{sec:eval}

To evaluate TripMD, we use the UAH-DriveSet \citep{romera_need_2016}, a publicly available naturalistic driving dataset including recorded trips from six different drivers that traveled in two specific routes in Madrid, Spain. The authors asked the volunteers to drive in these two routes mimicking three different driving behaviors - normal, aggressive and drowsy. Using their DriveSafe app \citep{bergasa_drivesafe:_2014, romera_real-time_2015}, the authors collected raw data from the accelerometer, GPS and camera of a smartphone mounted in the car and processed these signals to enrich the final dataset.

In the first experiment, we pick a single driver and explore in detail the outputs obtained from TripMD. Particularly, we do an exploratory analysis of the motifs extracted by TripMD and showcase the visualizations provided by our system.

In the second experiment, we focus on the task of identifying driving behaviors. We apply TripMD to the entire UAH-DriveSet. Then, using the known driving behaviors of all but one driver, we assign behavior scores to each motif cluster. Finally, we use those cluster behavior scores and the motifs extracted from the left-out driver to predict the behavior of each of that driver's trip.

In both experiments, we use the two-dimensional time-series of the lateral and longitudinal acceleration recordings. The recordings are already aligned with the correct car axis and denoised with a Kalman filter, which means we can use them directly. The data has a frequency of 10Hz, however, in order to speed computation and further reduce noise, we down-sample the time-series to a 5HZ frequency. Additionally, we use all the default parameters for TripMD as we found that they work well for this dataset.

The code to reproduce all the experiments can be consulted in our repository \footnote{\url{https://github.com/misilva73/tripMD}}.

\subsection{Analyzing a single driver with TripMD}

To showcase how our system can be used to explore the driving behavior of a single person, we run TripMD on the seven trips performed by one of the drivers in the UAH-DriveSet. This driver completed four trips in the secondary road route (two normal, one aggressive and one drowsy) and three trips in the motorway route (one for each driving behavior). 

The motif detection component found 281 motifs and, from these, 17 motifs were used to initialized the DTW-SOM model. These 17 motifs were the result of the pruning step presented in Section \ref{sec:tripmd}. The DTW-SOM model was initialized with a 5X5 grid since it is the smallest square grid that can contain the 17 pruned motifs. Then the 281 motifs were assigned to each DTW-SOM unit in the grid. DTW-SOM builds an optimal assignment by reducing the DTW distance between the units and their motif's centers. DTW-SOM also provides a visualization of the clusters in a two-dimensional grid (or network) that conserves the local similarity of the data. This means that two neighboring clusters in the two-dimensional network are similar. 

\begin{figure}[ht!]
    \centering
    \includegraphics[width=\linewidth]{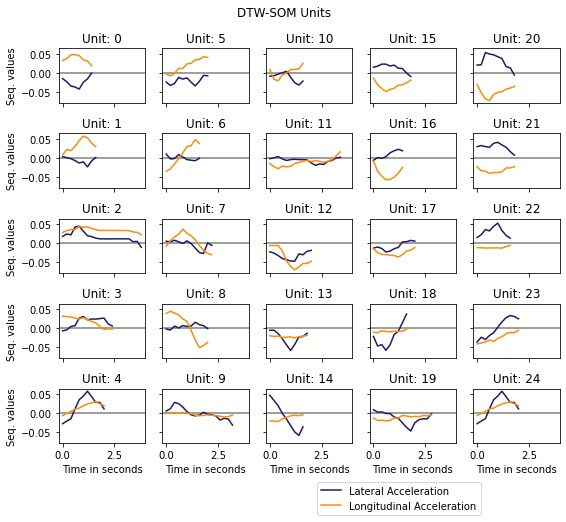}
    \caption{Lateral and longitudinal acceleration of the DTW-SOM's units. They were obtained by applying TripMD to the trips of a single driver from the UAH-DriveSet. Units are placed in the DTW-SOM two-dimensional grid.}
   \label{fig:single-units}
\end{figure}

Figure \ref{fig:single-units} shows the first visualizations provided by TripMD for the driver. It contains the lateral and longitudinal acceleration of each DTW-SOM unit, placed in the two-dimensional network. A unit here is a multi-dimensional subsequence that represents the cluster in a particular part of the DTW-SOM grid and thus this plot provides a summary of the main driving patterns extracted from the driver.

From this first chart, we can already see that TripMD is able of identifying a rich set of driving patterns, with lengths ranging from 1.5 to 3 seconds. It includes simple maneuvers, such as unit 22 that relates with a simple left turn without changes in longitudinal acceleration, and more complex maneuvers, for instance, unit 0 that corresponds to a right turn with acceleration.

Additionally, the grid maintains some local similarity, with adjacent units showing more similar acceleration patterns than units that are not adjacent. As an example, the neighboring units 15, 16, 20, and 21 all have similar driving patterns, with a clear brake maneuver and a slightly positive lateral acceleration.

\begin{figure}[ht!]
    \centering
    \includegraphics[width=0.5\linewidth]{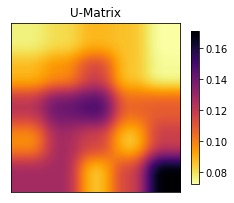}
    \includegraphics[width=0.4\linewidth]{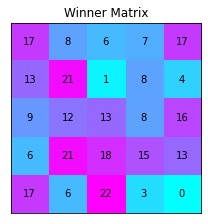}
    \caption{U-matrix and Winner Matrix of the DTW-SOM. The model was trained on the motifs extracted from the trips of a single driver from the UAH-DriveSet.}
   \label{fig:single-dtwsom-outputs}
\end{figure}

Figure \ref{fig:single-dtwsom-outputs} contains two additional visualizations provided by TripMD for the driver. These plots are classical ways of visualizing a SOM network and represent different information about each of the clusters arranged in the two-dimensional network. In both charts, the arrangement of the units in the two-dimensional grid is consistent to Figure \ref{fig:single-units}. 

The first chart is called U-Matrix and it shows how similar each unit is to its direct neighbors in the two-dimensional network, where the brighter the color, the closer a unit is to its neighbor. This visualization is helpful to understand where are the major groups of clusters within the network. For instance, the upper-right corner has a clearly defined groups of four cluster that are very similar, which corresponds to the units 15, 16, 20 and 21 with the brake maneuver discussed above.

The second chart is called Winner Matrix and it provides information about the cluster size of each unit. Particularly, it displays the exact number of motifs in each cluster on top of corresponding unit. This plot can be used to gauge how relevant each driving pattern is. For instance, unit 24 in the lower right corner contains no motifs, which means that this pattern is not needed to summarize the driver's behavior.

\begin{figure}[ht!]
    \centering
    \includegraphics[width=0.95\linewidth]{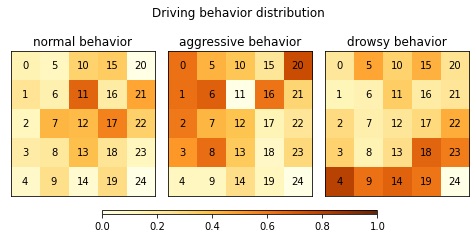}
    \caption{Driving behavior rates for each DTW-SOM unit. For each unit, it shows the percentage of motif subsequences that come from each driving behavior.}
   \label{fig:single-behavior}
\end{figure}

Besides these default TripMD plots, Figure \ref{fig:single-behavior} contains information about the distribution of each driving behavior in the clusters extracted by our system. For each driving behavior, we compute the number of motif subsequences from the trips with that behavior that belong to each DTW-SOM cluster. Then, we divide each cluster count by the total number of motif subsequences for all the driver's trips that belong to that cluster to achieve the rate presented in the plots. So, for instance, 80\% of the motif subsequences associated to motifs that belong to the cluster 4 come from trips with a drowsy behavior.

Interestingly, we can see that the three driving behaviors have very different distributions of their motif subsequences among the clusters. Clusters 11 and 17 have a clear majority of subsequences from normal trips and these clusters relate to a "no maneuver" pattern and a soft brake, respectively.

The aggressive trips cover a higher variety of patterns, with 7 clusters showing a clear majority of subsequences from these trips. This increase in representation is expected as more motif subsequences will be extracted from trips where the driver performs more maneuvers. Most of these 7 clusters contain sharp acceleration patterns, which is usually associated with aggressive driving. Examples are the right turn with a pronounced brake in unit 20, the brake-acceleration pattern in unit 6 and the quick acceleration in unit 1. These sharp acceleration maneuvers without lateral movements are specially telling as they are associated with tailgating behavior, which in turn is a classical aggressive driving behavior.

Finally, the clusters with higher rates of subsequences coming from drowsy trips are located in the lower row of the DTW-SOM grid, excluding unit 24. Units 14 and 18 contain a drift pattern, which is made of two consecutive lateral movements in opposing sides. It is very interesting to see these patterns here as they are usually present in cases where a tired driver lets the car deviate from a lane and then quickly recovers with a sharp turn.

\subsection{Identifying driving behavior with TripMD}

From the first experiment, we see that TripMD can summarize the trips from a single driver so that different driving behaviors can be identified. However, to further test our system, we focus on a harder task, namely, identifying the driving behavior of an unknown driver from a set of drivers whose behavior we know.

To accomplish this, we apply TripMD to the entire UAH-DriveSet and retrieve the main driving patterns of all those trips. Then, using the known driving behavior of five drivers (the training drivers), we derive scores for all the trips of the remaining driver (the testing driver). This testing driver was the same used in the first experiment. For each trip of the testing driver, a single score is computed for each of the three behaviors - normal, aggressive and drowsy. To compute the score for a specific testing trip and a given behavior $b$, we use the following process:

\begin{enumerate}[noitemsep]
    \item For each DWT-SOM cluster $c_i$:
    \begin{enumerate}[noitemsep]
        \item Compute the rate $r_i^b = \frac{n_{i}^b}{n_{i}}$, where $n_{i}^b$ is the number of subsequences in cluster $c_i$ that belong to training trips of the behavior $b$ and $n_{i}$ is the total number of subsequences in cluster $c_i$ that belong to training trips.
        \item Compute $\hat{n_i}$, which is the number of motif subsequences in cluster $c_i$ that belong to testing trip.
        \item Derive the behavior score of cluster $c_i$ as $s_i^b = r_i^b \times \hat{n_i}$.
    \end{enumerate}
    \item Compute the trip's behavior score as $s^b = \sum_{i=1}^k s_i^b$, where k is the number of DWT-SOM clusters.
\end{enumerate}

After computing the three behavior scores for a testing trip, its predicted behavior is simply the behavior with the highest score. Finally, we compare the predicted behavior of each testing driver's trips with the real behavior performed in those trips. Table \ref{tab:behavior-scores} summarizes the results. 

\begin{table}[!ht]
\centering
\footnotesize
\begin{tabular}{@{}lcccc@{}}
\toprule
\multicolumn{1}{c}{\textbf{Route}} & \textbf{Behavior} & \begin{tabular}[c]{@{}c@{}}\textbf{Aggressive}\\ \textbf{score}\end{tabular} & \begin{tabular}[c]{@{}c@{}}\textbf{Drowsy}\\ \textbf{score}\end{tabular} & \begin{tabular}[c]{@{}c@{}}\textbf{Normal}\\ \textbf{score}\end{tabular} \\ \midrule
Motorway & Normal & 94.1 & 78.3 & \textbf{95.5} \\
Secondary & Normal & 35.5 & 41.1 & \textbf{43.3} \\
Secondary & Normal & 47.0 & 35.9 & \textbf{48.1} \\\midrule
Motorway & Aggressive & \textbf{103.1} & 72.6 & 96.3 \\
Secondary & Aggressive & \textbf{91.0} & 52.2 & 77.8 \\\midrule
Motorway & Drowsy & 97.5 & \textbf{137.9} & 120.6 \\
Secondary & Drowsy & 69.7 & 81.9 & \textbf{82.4} \\ \bottomrule
\end{tabular}
\caption{Behavior scores for each testing trip. The score of the predicted behavior for each trip is highlighted in bold. The column named \textit{Behavior} contains the real behavior of the trip.}
\label{tab:behavior-scores}
\end{table}

The testing driver contains seven trips and, from these, TripMD assigns the correct behavior to all but one trip. The trip where the predicted behavior does not match the real behavior is the drowsy trip of the secondary road.

\begin{figure}[ht!]
    \centering
    \includegraphics[width=0.95\linewidth]{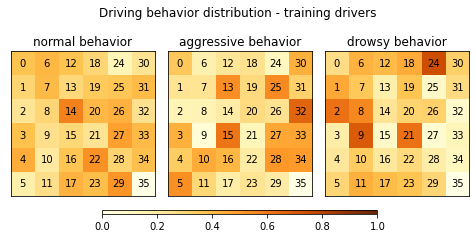}
    \includegraphics[width=0.95\linewidth]{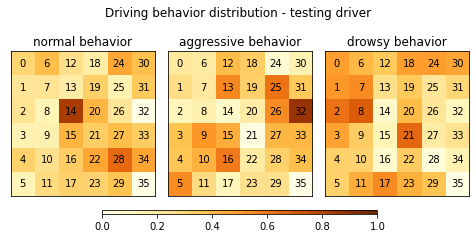}
    \caption{Behavior rates for the DTW-SOM units. For each unit, it shows the percentage of motif subsequences that come from each driving behavior. The three plots at the top contain the rates of the training trips while the three plots at the bottom contain the rates of the testing trips.}
   \label{fig:all-behavior}
\end{figure}

To further illustrate these results, Figure \ref{fig:all-behavior} shows the behavior rates of the DTW-SOM clusters for the training drivers and the testing driver. Here we can observe that the behavior distributions of the training drivers are close to behavior distributions of the testing driver. For instance, units 14 has a high normal behavior rate in both sets of trips, units 13, 25 and 32 have similarly high aggressive behavior rates and units 2, 8 and 21 have high drowsy behavior rates for both the testing and training drivers.

Going back to the scores in Table \ref{tab:behavior-scores}, we note that the trips with the normal behavior and the drowsy trip on the secondary road route have two predicted behaviors with very close scores, which may indicate that TripMD is not consistently catching these particular behaviors well.

To further explore this issue, we use a resampling method to measure the stability of the computed scores. Concretely, we apply random sampling with replacement to the list of motifs extracted by TripMD and update the DTW-SOM clusters. The DTW-SOM is not retrained, and thus its units do not change. Instead, only the group of motifs assigned to each cluster are sampled. Since we take a sample of the same size as the original list of motifs, some motifs are repeated multiple times while others are never sampled. 

\begin{figure}[ht!]
    \centering
    \includegraphics[width=0.9\linewidth]{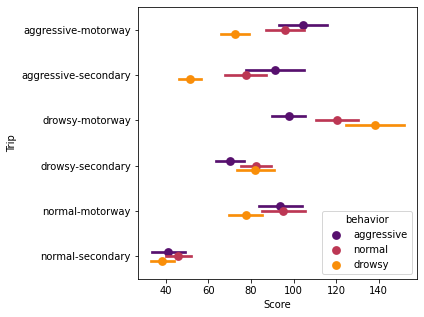}
    \caption{Summary statistics of the each testing trip's behavior scores obtained with 1000 samples with replacement. The points encode the average scores while the errors bars encode the standard deviation.}
   \label{fig:boot-analysis}
\end{figure}

With the new sampled motifs, we then apply the same procedure to compute the behavior score of all the trips performed by the testing driver. This sampling method can be repeated many times and, for each sample, we get a new estimation of each trip's behavior scores. Figure \ref{fig:boot-analysis} shows the average behavior scores and their standard deviations for the testing driver's trips obtained after 1000 samples.

As expected, in the drowsy trip of the secondary road, the distribution of the scores for the normal and drowsy behavior are very close. Similarly, some scores for the normal trips in both the secondary and motorway routes overlap, which agrees with the initial assessment that TripMD is not strong at identifying the trips with the normal behaviors and the drowsy trip on the secondary road.

On the other hand, the aggressive trips and the drowsy trip on the motorway have the correct behavior score more consistently higher than the remaining behavior scores. Thus, in these cases, TripMD is identifying successfully the underlying behavior of the driver.

\section{Conclusion}
\label{sec:conclusion}

In this paper, we propose a system called TripMD, which identifies the main driving patterns from sensor recordings such as acceleration and velocity. Compared to previous work, our system not only extracts the time-series patterns present in trips recordings but is also capable of summarizing those patterns in a space-efficient visualization. This feature is highlighted in our first experiment, where we demonstrate that TripMD can discover a wide range of driving patterns from the trips performed by a single driver of the UAH-DriveSet dataset. We also conclude that the three driving behaviors marked in the dataset (normal, aggressive and drowsy) have distinct distributions among the extracted driving patterns, which can be used to determine the driving behavior of a driver from the behaviors of other drivers.

Even though the results seem promising, there are still areas of improvement. Firstly, because of the Variable SAX discretization, the motif detection algorithm used in TripMD is not an \textit{exact} algorithm. This means that we cannot guarantee to extract all the variable-length motifs. There are some new motif detection algorithms that claim to be exact, however, we could not find one that was capable of extracting motifs with member's subsequences of difference sizes. Thus, investigating exact motif detection algorithms that work with variable-length motifs could be an interesting line for improvement.

Additionally, we should further test TripMD with more datasets and different tasks, such as understanding whether TripMD can be used to distinguish between drivers with prior accidents from drivers without accidents and to inform car insurance pricing models. Testing TripMD with different datasets would also be helpful to further validate and fine-tune the default parameters of the system such as the default letter size. Another part that could be further tested with other datasets is the choice of percentiles for the VSAX representation. 

\section*{Acknowledgments}

This research did not receive any specific grant from funding agencies in the public, commercial, or not-for-profit sectors.

\bibliographystyle{elsarticle-num-names} 
\bibliography{references.bib}

\end{document}